\documentclass[conference, 11pt]{IEEEtran}
\usepackage[top=0.75in, left=0.68in, right=0.68in, bottom=1.05in]{geometry}
\usepackage{graphicx}
\usepackage{cite}
\usepackage{amsmath,amssymb,amsfonts}
\usepackage{algorithmic}
\usepackage{graphicx}
\usepackage{textcomp}
\usepackage{xcolor}
\usepackage{graphicx}
\usepackage{amsmath}
\usepackage{array}
\usepackage{multirow}
\usepackage{booktabs}
\usepackage{nohyperref} 
\usepackage{enumitem}
\def\BibTeX{{\rm B\kern-.05em{\sc i\kern-.025em b}\kern-.08em
    T\kern-.1667em\lower.7ex\hbox{E}\kern-.125emX}}
\begin{document}

\title{TinyLLM: Evaluation and Optimization of Small Language Models for Agentic Tasks on Edge Devices
}


\author{
\IEEEauthorblockN{
Mohd Ariful Haque\textsuperscript{1},
Fahad Rahman\textsuperscript{2},
Kishor Datta Gupta\textsuperscript{3}, 
Khalil Shujaee\textsuperscript{4},
Roy George\textsuperscript{5}}

\IEEEauthorblockA{
Department of Cyber-Physical Systems, 
Clark Atlanta University, USA\textsuperscript{1,3,4,5}\\
Department of Computer Science and Engineering, United International University, Bangladesh\textsuperscript{4}}

\texttt{mohdariful.haque@students.cau.edu}$^{1}$,
\texttt{frahman203014@bscse.uiu.ac.bd}$^{2}$, \\
\texttt{kgupta@cau.edu}$^{3}$,
\texttt{kshujaee@cau.edu}$^{4}$,
\texttt{rgeorge@cau.edu}$^{5}$
}

\maketitle

\begin{abstract}
This paper investigates the effectiveness of small language models (SLMs) for agentic tasks (function/tool/API calling) with a focus on running agents on edge devices without reliance on cloud infrastructure. We evaluate SLMs using the Berkeley Function Calling Leaderboard (BFCL) framework and describe parameter-driven optimization strategies that include supervised fine-tuning (SFT), parameter-efficient fine-tuning (PEFT), reinforcement learning (RL)-based optimization, preference alignment via Direct Preference Optimization (DPO), and hybrid methods. We report results for models including TinyAgent, TinyLlama, Qwen, and xLAM across BFCL categories (simple, multiple, parallel, parallel-multiple, and relevance detection), both in live and non-live settings, and in multi-turn evaluations. We additionally detail a DPO training pipeline constructed from AgentBank data (e.g., ALFRED), including our conversion of SFT data to chosen-rejected pairs using TinyLlama responses as rejected outputs and manual validation. Our results demonstrate clear accuracy differences across model scales where medium-sized models (1-3B parameters) significantly outperform ultra-compact models (<1B parameters), achieving up to 65.74\% overall accuracy, and 55.62\% multi-turn accuracy with hybrid optimization. This study highlights the importance of hybrid optimization strategies that enable small language models to deliver accurate, efficient, and stable agentic AI on edge devices, making privacy-preserving, low-latency autonomous agents practical beyond the cloud.
\end{abstract}


\section{Introduction}
The advancement of large language models has enabled autonomous agents to interact naturally with tools, APIs, and dynamic environments. However, typical agentic applications rely on models exceeding 7B parameters and cloud infrastructure, which introduces 1) increased inference latency due to network round-trips, 2) a constant requirement for internet connectivity, 3) elevated risks to user privacy and data security, 4) ongoing cloud computing costs, and 5) high GPU costs. These issues limit the practicality and scalability of deploying autonomous agents in latency-sensitive, privacy-critical, or resource-constrained environments.

This research addresses the gap in understanding Small language models (SLMs) with under 3 billion parameters for agentic tasks on edge devices. Using the Berkeley Function Calling Leaderboard (BFCL) \cite{bfcl2025v4} framework, it benchmarks SLMs like TinyAgent \cite{erdogan2024tinyagentfunctioncallingedge}, TinyLlama \cite{zhang2024tinyllama}, Qwen \cite{yang2024qwen2}, and xLAM \cite{zhang2024xlamfamilylargeaction} across diverse multi-turn, live, and AST-level function-calling scenarios. The study explores optimization techniques such as supervised fine-tuning, parameter-efficient tuning, reinforcement learning, direct preference optimization (DPO) \cite{rafailov2023dpo}, and hybrid approaches for robust, efficient edge deployment.

Three main contributions are made: a comprehensive benchmark revealing how model size and task complexity affect edge performance; a practical DPO-based pipeline converting supervised training data into preference pairs for efficient model alignment without costly reinforcement learning; and an analysis of latency, accuracy, and deployment trade-offs to guide creation of fast, private, and effective autonomous agents that operate beyond cloud reliance. This work offers a roadmap for leveraging SLMs in real-world, resource-constrained environments.

\section{Objectives}
\begin{itemize}[leftmargin=*,noitemsep,topsep=0pt]
  \item Demonstrate the viability of SLMs (smaller than 3B parameters) for agentic tasks: tool/API calling.
  \item Replicate and benchmark TinyAgent using its codebase/evaluation and train an SLM that surpasses it, to establish a baseline.
  \item Iteratively improve and compare models, logging performance across variants to yield a candidate suitable for practical edge deployment.
\end{itemize}

\section{Background and Literature Review}
\label{sec:litreview}
This section surveys four strands of prior work related to our study: (i) tool-augmented language models and agent paradigms, (ii) benchmarks and leaderboards for tool/function calling, (iii) small language models (SLMs) designed for efficiency and edge deployment, and (iv) optimization and alignment techniques that make LLMs/SLMs reliable agents.

\subsection{Tool-Augmented LLMs and Agent Paradigms}

A substantial body of work studies how language models can \emph{use tools}—APIs, functions, databases, and external programs—to improve correctness, reduce hallucination, and perform decision-making in interactive environments. Toolformer showed that LMs can self-supervise creation of API-calling data and learn \emph{when/what/how} to call tools during generation \cite{schick2023toolformer}. The paper presents ATLASS, a closed-loop framework that enables LLM agents to dynamically generate, select, and refine tools on demand, overcoming limitations of fixed toolsets for more adaptive, complex problem-solving \cite{ariful2025atlass}. Gorilla proposed fine-tuning strategies and retrieval over API documentation to robustly generate API invocations and generalize across changing specs \cite{patil2023gorilla}. Beyond tool-calling, a complementary line of research frames LLMs explicitly as agents that reason and act. ReAct interleaves reasoning traces (``thought'') with action selection, improving tool-use reliability and exploration \cite{yao2023react}. Reflection equips agents with self-reflective feedback loops and episodic memory, enabling iterative self-improvement without weight updates \cite{shinn2023reflexion}. In embodied, open-ended settings, Voyager demonstrates automatic skill acquisition and curriculum building via iterative program synthesis in Minecraft \cite{wang2023voyager}. These paradigms collectively motivate precise function calling and multi-stage decision-making under constraints similar to edge scenarios.

\subsection{Benchmarks and Leaderboards for Function/Tool Calling}
Multiple benchmarks standardize evaluation for tool use. API-Bank provides a runnable evaluation and diverse API ecosystem to assess LLMs' tool-use capabilities \cite{li2023apibank}. ToolLLM/ToolBench further introduces large-scale instruction-tuning data for real-world APIs, enabling systematic training and analysis \cite{qin2023toollm}. StableToolBench focuses on \emph{stable}, reproducible tool-use evaluation by simulating APIs and addressing variance due to external services \cite{guo2024stabletoolbench}. For end-to-end agent performance, AgentBench integrates multi-environment, multi-skill tasks to evaluate reasoning and decision-making in interactive settings \cite{liu2024agentbench}. Closer to our study, the Berkeley Function Calling Leaderboard (BFCL) establishes a comprehensive, frequently updated obstacle course for function/tool invocation across simple, parallel, abstention, and multi-turn scenarios, reporting AST correctness and executability \cite{bfcl2025v4}. Complementarily, T-Eval decomposes tool utilization into sub-processes (instruction following, planning, reasoning, retrieval, understanding, and review), yielding fine-grained diagnostics beyond outcome-only scoring \cite{chen2024teval}. Together, these resources frame the evaluation axes we adopt and extend for SLMs.

\subsection{Small Language Models for Efficient and Edge Deployment}
Recent SLMs target strong capability-per-FLOP with compact parameter counts. TinyLlama (1.1B) demonstrates that careful pretraining (e.g., data curation, training efficiency improvements) can yield competitive performance across downstream tasks despite the small footprint suitable for edge devices such as smartphones and IoT gateways \cite{zhang2024tinyllama}. The Qwen series reports families of open-weight models spanning sub-3B scales up to tens of billions, with instruction-tuned and specialized variants (e.g., coding), offering practical bases for on-device function calling and tool use \cite{yang2024qwen2}. These families are pertinent to our goal of enabling agentic capabilities on constrained hardware without sacrificing too much accuracy.

\subsection{Optimization and Alignment for Agentic Reliability}
Optimizing models for tool use and multi-step decision-making draws on both efficient fine-tuning and preference-aligned reinforcement learning. LoRA injects low-rank adapters to dramatically cut trainable parameters while preserving quality \cite{hu2021lora}; QLoRA extends this to 4-bit quantized backbones to push memory usage into single-accelerator regimes \cite{dettmers2023qlora}. RLHF first established a practical pipeline for aligning models to human preferences via supervised fine-tuning plus reinforcement learning \cite{ouyang2022instructgpt}, often instantiated with Proximal Policy Optimization (PPO) \cite{schulman2017ppo}. More recently, Direct Preference Optimization (DPO) replaces policy-gradient RL with a stable classification-style objective that directly optimizes against preference pairs with strong empirical performance \cite{rafailov2023dpo}. In conjunction, these methods provide the knobs we exploit in our parameter-driven optimization of SLM agents, balancing compute, data quality, and robustness.

\paragraph{Summary.} Prior work establishes (1) strong tool-use and agent paradigms that benefit from accurate function calls \cite{schick2023toolformer,patil2023gorilla,yao2023react,shinn2023reflexion,wang2023voyager}; (2) rigorous, reproducible evaluations for tool calling and agent behavior \cite{li2023apibank,qin2023toollm,guo2024stabletoolbench,liu2024agentbench,bfcl2025v4,chen2024teval}; and (3) optimization recipes that make small models viable on constrained hardware \cite{hu2021lora,dettmers2023qlora,ouyang2022instructgpt,schulman2017ppo,rafailov2023dpo}. Our study positions SLMs within this ecosystem and focuses specifically on their function-calling competence and agentic reliability under edge-device constraints.

\section{Methodology}
Our pipeline (Fig. \ref{fig:proposed-pipeline}) comprises four stages: data processing, model selection, fine-tuning techniques, and model evaluation \& comparison.

\subsection{Data Processing}
We acquire diverse agentic datasets and manually analyze and evaluate them, filtering out low-quality data while preserving some failed/low-quality trajectories as negative signals to improve robustness. High-quality data passes directly into training; filtered sets are routed through a data utilization pipeline that attempts to convert portions into usable training signals. We merge the high-quality and converted sets into the final training dataset.

\subsection{Model Selection}
We consider SLMs such as Qwen, TinyLlama, Falcon, Gemma, and Phi, with a preference for models under 3B parameters. Before selection, candidate bases are evaluated using the BFCL framework; a base model that already performs well on agentic tasks is more likely to improve after training. The process is iterative---we revisit model selection after each training--evaluation cycle.

\subsection{Fine-Tuning Techniques and Training Setup}
We performed fine-tuning using multiple datasets, conducting both PEFT (e.g., LoRA/QLoRA) and full-parameter fine-tuning. Initial supervised fine-tuning (SFT) results were not particularly impressive, motivating exploration of preference-based and RL-based methods. Results were visualized in internal figures and spreadsheets (referenced in our working notes).

\section{Evaluation Framework and Metrics}\label{sec:framework}
\subsection{Framework Selection: T-Eval vs.\ BFCL}
While T-Eval decomposes agentic capability into sub-processes (planning, reasoning, instruction following), it requires a unique \texttt{meta\_template.py} per model. To streamline large-scale comparisons across diverse tokenizers/architectures, we employ BFCL, which enables \textit{plug-and-play} evaluation by specifying a model name and receiving standardized outputs.

\subsection{BFCL Categories}
BFCL assesses function-calling ability across:
\begin{enumerate}[leftmargin=*,noitemsep,topsep=0pt]
  \item \textbf{Simple Function:} Single-function documentation; generate a correct call.
  \item \textbf{Multiple Function:} Choose among 2--4 candidate functions and produce a valid call.
  \item \textbf{Parallel Function:} Invoke the same function multiple times from one query.
  \item \textbf{Parallel Multiple Function:} For multiple functions, determine required invocations and supply calls.
  \item \textbf{Relevance Detection:} Abstain when no provided function is appropriate.
\end{enumerate}

\begin{figure}[t]
  \centering
  \includegraphics[width=\linewidth]{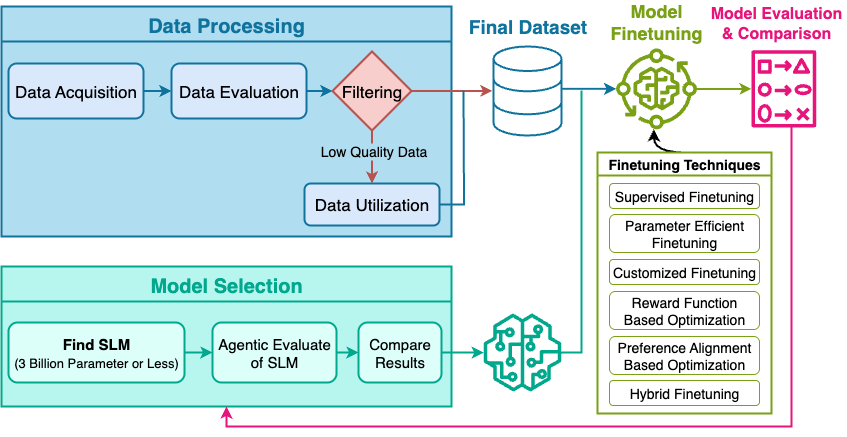}
  \caption{Proposed Pipeline of TinyLLM}
  \label{fig:proposed-pipeline}
\end{figure}

\subsection{Metrics}
We report: (i) Abstract Syntax Tree (AST) accuracy (syntactic correctness), (ii) Executable evaluation (semantic correctness when feasible), and (iii) Overall accuracy (unweighted average across subcategories). BFCL also distinguishes prompt-based vs.\ native function-calling models, includes multi-turn metrics, and supports reproducible benchmarking with transparent outputs, as well as operational considerations (e.g., cost, latency).

\section{Dataset Composition}\label{sec:dataset}
We use the Gorilla OpenFunctions dataset, expanded from 100 to 2{,}000 data points with broader coverage and quality. It spans 40 sub-domains (e.g., Mathematics/Algebra, Sports/Soccer, Finance/Mortgage) and contains real-world examples across websites and multiple programming languages, enabling assessment beyond computing/cloud into sports and law.

\begin{figure}[t]
  \centering
  \includegraphics[width=\linewidth]{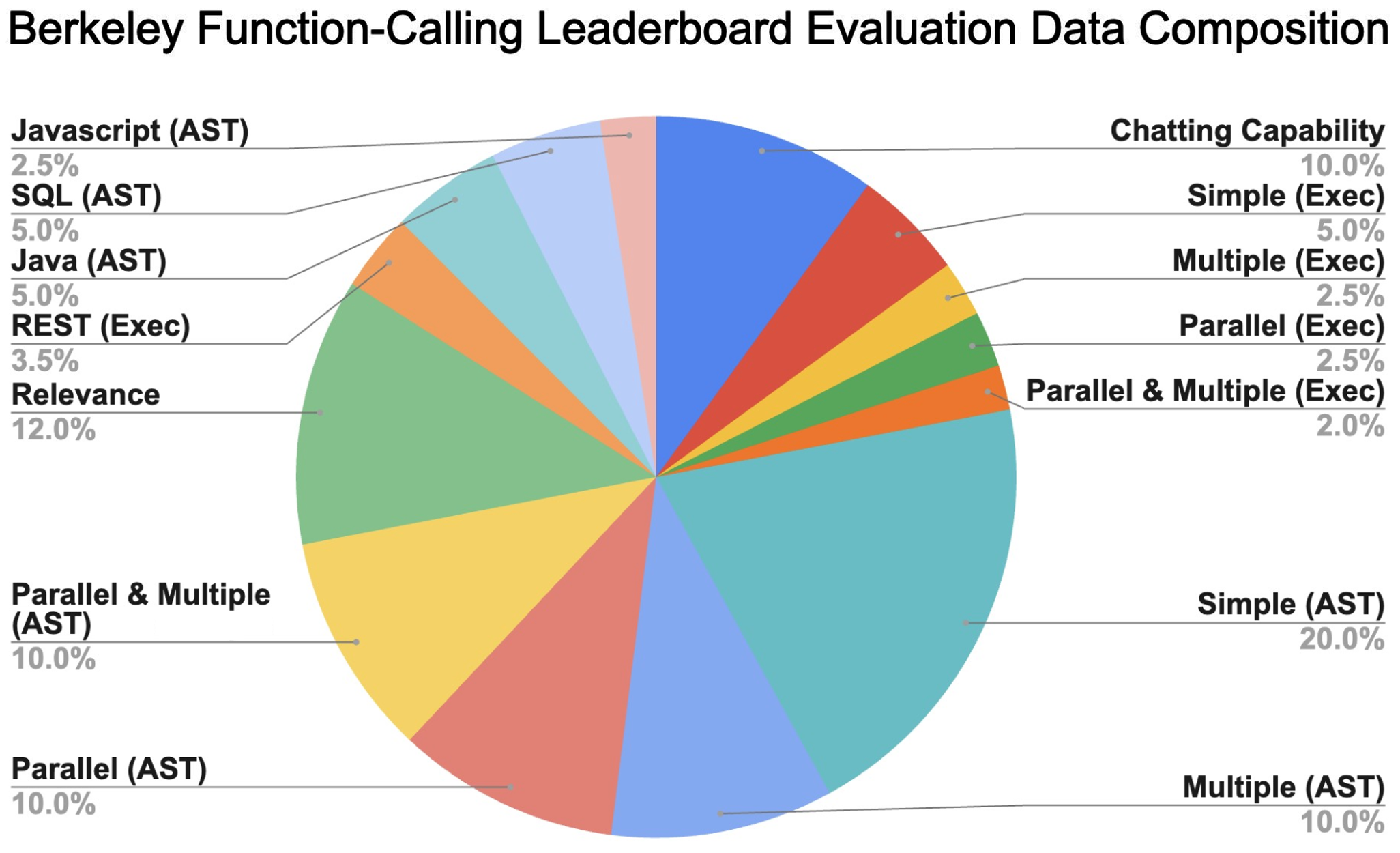}
  \caption{Evaluation dataset's data distribution}
  \label{fig:dataset-distribution}
\end{figure}

Dataset (ref \ref{fig:dataset-distribution} is composed of multiple categories that reflect different levels of complexity and evaluation modalities. The largest segment corresponds to Simple (AST) tasks, which constitute 20\% of the dataset and emphasize syntactic correctness in generating single function calls. Relevance detection forms 12\%, highlighting the ability of models to abstain when no appropriate function exists, an essential safeguard for trustworthy deployment. Chatting capability accounts for 10\%, representing the importance of conversational context in tool invocation. Similarly, 10\% each is allocated to Parallel (AST), Parallel and Multiple (AST), and Multiple (AST), which collectively stress complex scenarios requiring either multiple simultaneous invocations or selection among multiple candidate functions. Execution-level evaluation occupies a smaller fraction: Simple (Exec) represents 5\%, while Multiple (Exec), Parallel (Exec), and Parallel and Multiple (Exec) together contribute about 7\%, reflecting the additional cost of execution testing. Cross-language and API diversity is captured by SQL (AST) at 5\%, Java (AST) at 5\%, REST (Exec) at 3.5\%, and JavaScript (AST) at 2.5\%. This composition reveals that the dataset is heavily weighted toward AST-based evaluation, with approximately 60\% dedicated to syntactic correctness across simple and complex tasks, while also integrating execution-based validation, multilingual support, and relevance and conversation-specific categories. The balance ensures that models are tested not only for their ability to generate syntactically valid calls, but also for safe abstention, conversational integration, cross-language robustness, and practical executability.

\section{Parameter-Driven Optimization of LLM-based Agents}\label{sec:opt}
We focus on optimizing LLM parameters for autonomous agents (decision-making, multi-step reasoning, dynamic interaction). We group methods into three approaches.

\subsection{Conventional Fine-Tuning-based Optimization}
\textbf{Trajectory data construction} comprises expert-annotated data, strong-LLM-generated trajectories, self-exploration, and multi-agent collaboration. Data filtering via environment rewards, human/rule-based filtering, or model-based evaluation retains quality trajectories. \textbf{Fine-tuning techniques} include SFT (full/partial), PEFT (LoRA/QLoRA), and customized task-specific strategies. \textbf{Use of low-quality data} (failed trajectories) can improve robustness via negative feedback. \textbf{Summary:} Effective alignment to agent tasks, but dependent on data quality, with overfitting risks and limited adaptability during interaction.

\subsection{Reinforcement Learning-based Optimization}
\textbf{Reward function-based optimization} employs PPO/actor--critic with explicit rewards (environmental, model-based, or composite). Challenges include reward design and compute cost. \textbf{Preference alignment-based optimization} (e.g., DPO) uses pairwise preference data (better/worse responses) without environment interaction, simplifying training and improving alignment while depending on preference data diversity/quality. \textbf{Summary:} RL enhances adaptability/robustness; reward-based excels in complex interactions, preference-based is more stable/sample-efficient but data-limited.

\subsection{Hybrid Fine-Tuning Optimization}
A sequential hybrid strategy typically initializes with SFT on expert data and refines with RL (PPO/DPO variants). Some methods iterate SFT--RL cycles or incorporate offline critics/cyclic fine-tuning on updated trajectory pairs. Examples include ReFT, AgentGym, ETO, Re-ReST, and AGILE. \textbf{Summary:} Hybrids balance stability (SFT) and adaptability (RL) but increase compute cost and require careful design for integration.

\subsection{Overall Summary of Optimization Methods}
Parameter-driven optimization for agentic SLMs refines parameters via curated trajectories, RL from interaction/rewards/preferences, and hybrid pipelines. Key challenges include data quality, dynamic adaptation, and compute efficiency.

\section{DPO Training with AgentBank}
We explored RLHF approaches and selected DPO for its simplicity and potential. High-quality chosen--rejected pairs are required; we selected the AgentBank dataset and converted existing SFT data: the AgentBank response serves as the chosen entity, and rejected responses are generated by TinyLlama/TinyLlama-1.1B-Chat-v1.0. Manual validation confirmed the rejected outputs are consistently lower quality. We completed conversion of the ALFRED dataset from AgentBank and successfully performed DPO training; work on ALFWorld is in progress. Upon full dataset preparation, we will proceed with large-scale DPO training and evaluate using the Gorilla BFCL framework.

\begin{figure}[t]
  \centering
  \includegraphics[width=\linewidth]{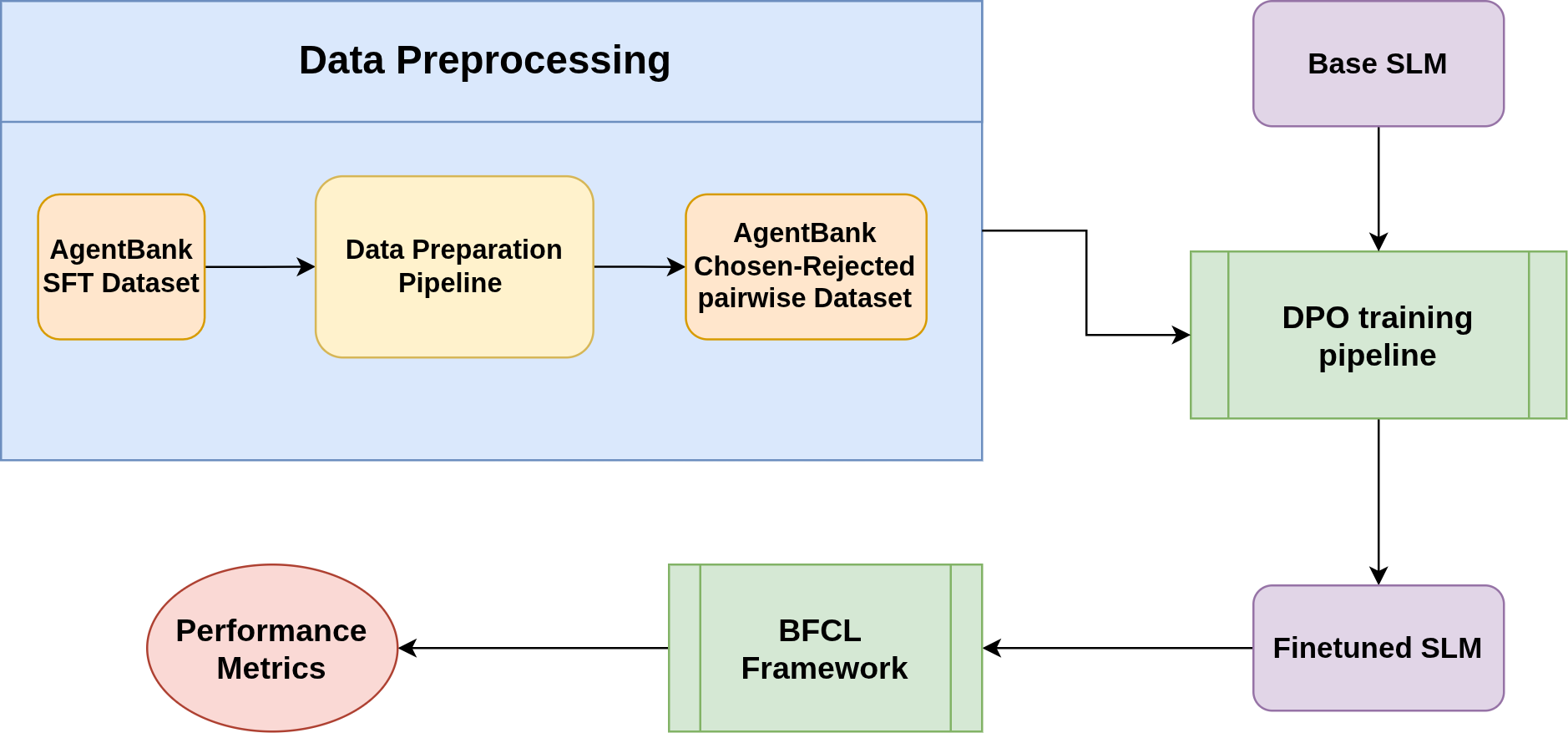}
  \caption{Direct Preference Optimization Pipeline}
  \label{fig:proposed-pipeline}
\end{figure}

\section{Results and Analysis}
We compare SLMs on BFCL across live/non-live (AST-level) and multi-turn settings.

\subsection{Overall Performance (BFCL)}
\begin{table*}[ht]
\centering
\caption{Overall Accuracy (BFCL)}
\begin{tabular}{lcccc}
\toprule
\textbf{Model} & \textbf{Overall Acc} & \textbf{Live} & \textbf{Non-live} & \textbf{Multi-turn} \\
\midrule
xLAM-2-3b-fc-r (FC)     & 65.74\% & 81.03\% & 88.22\% & 55.62\% \\
Qwen3-4B (Prompt)       & 62.04\% & 75.52\% & 82.58\% & 35.25\% \\
Qwen3-1.7B (Prompt)     & 55.49\% & 63.48\% & 80.03\% & 16.88\% \\
xLAM-2-1b-fc-r (FC)     & 53.97\% & 61.57\% & 72.42\% & 8.38\%  \\
Qwen3-0.6B (Prompt)     & 45.76\% & 58.86\% & 67.78\% & 1.38\%  \\
TinyLlama-1.1B-32k-Inst & 19.73\% & 39.18\% & 20.00\% & 0.00\%  \\
TinyAgent-1.1B          & 19.70\% & 39.09\% & 20.00\% & 0.00\%  \\
\bottomrule
\end{tabular}
\label{tab:overall}
\end{table*}
The results in Table~\ref{tab:overall} summarize the performance of several small and medium-scale language models on the Berkeley Function-Calling Leaderboard. Among the evaluated systems, \textit{xLAM-2-3b-fc-r (FC)} achieves the strongest results, reaching 65.74\% overall accuracy, with 81.03\% in live execution, 88.22\% in non-live AST-level evaluation, and 55.62\% in multi-turn interactions. The \textit{Qwen3-4B (Prompt)} model follows closely with an overall accuracy of 62.04\%, demonstrating solid performance in live (75.52\%) and non-live (82.58\%) settings, although its multi-turn capability drops to 35.25\%. The smaller \textit{Qwen3-1.7B (Prompt)} and \textit{xLAM-2-1b-fc-r (FC)} models yield moderate results, with overall accuracies of 55.49\% and 53.97\%, respectively, while their multi-turn accuracies fall to 16.88\% and 8.38\%. The lightweight \textit{Qwen3-0.6B (Prompt)} performs at 45.76\% overall accuracy, but its multi-turn accuracy is almost negligible at 1.38\%. In contrast, the ultra-compact models \textit{TinyLlama-1.1B-32k-Instruct} and \textit{TinyAgent-1.1B} exhibit very weak performance, both achieving only about 19.7\% overall accuracy, with non-live accuracy plateauing at 20\% and no success at all on multi-turn tasks. These findings highlight a clear performance hierarchy: larger SLMs such as xLAM and Qwen are significantly stronger across all categories, particularly in complex multi-turn evaluations, whereas extremely compact models like TinyLlama and TinyAgent remain inadequate for agentic scenarios that require robust reasoning and execution.

\subsection{Live Function Calling (Real-world Execution)}
\begin{table*}[ht]
\centering
\caption{Live Accuracy (as recorded)}
\begin{tabular}{lccccccc}
\toprule
\textbf{Metric} & \textbf{Q3-4B} & \textbf{Q3-1.7B} & \textbf{Q3-0.6B} & \textbf{x2-3B} & \textbf{x2-1B} & \textbf{TL-1.1B} & \textbf{TA-1.1B} \\
\midrule
Non\_Live Overall Acc & 81.03\% & 75.52\% & 63.48\% & 61.57\% & 58.86\% & 39.18\% & 39.09\% \\
AST Summary            & 80.38\% & 70.84\% & 52.26\% & 63.36\% & 54.63\% & 0.00\%  & 0.07\%  \\
Simple AST             & 84.50\% & 75.19\% & 60.47\% & 74.03\% & 67.83\% & 0.00\%  & 0.39\%  \\
Python Simple AST      & 79.39\% & 69.90\% & 50.14\% & 61.06\% & 52.23\% & 0.00\%  & 0.00\%  \\
Python Simple AST      & 81.25\% & 56.25\% & 62.50\% & 62.50\% & 43.75\% & 0.00\%  & 0.00\%  \\
Python Simple AST      & 79.17\% & 75.00\% & 50.00\% & 50.00\% & 25.00\% & 0.00\%  & 0.00\%  \\
Multiple AST           & 82.09\% & 83.33\% & 80.84\% & 58.28\% & 64.63\% & 100.00\%& 99.66\% \\
Parallel AST           & 77.78\% & 44.44\% & 55.56\% & 88.89\% & 94.44\% & 0.00\%  & 0.00\%  \\
Parallel Multiple AST  & 81.03\% & 75.52\% & 63.48\% & 61.57\% & 58.86\% & 39.18\% & 39.09\% \\
Irrelevance Detection  & 80.38\% & 70.84\% & 52.26\% & 63.36\% & 54.63\% & 0.00\%  & 0.07\%  \\
\bottomrule
\end{tabular}
\\[3pt]
\footnotesize Q3: Qwen3; x2: xLAM-2; TL: TinyLlama; TA: TinyAgent.
\label{tab:live}
\end{table*}
Table~\ref{tab:live} presents the live execution accuracy across models, where correctness is determined by running the generated calls and verifying their outputs. The results confirm that larger small-scale models maintain strong execution reliability. For instance, \textit{xLAM-2-3b-fc-r (FC)} achieves 81.03\% non-live overall accuracy with 63.36\% AST summary and robust performance across multiple categories, including 74.03\% for Simple AST and 88.89\% for Parallel AST. The \textit{Qwen3-4B (Prompt)} model shows similar strength with 75.52\% non-live accuracy and 84.50\% Simple AST performance, although its execution performance is more variable in multi-function scenarios. The \textit{Qwen3-1.7B (Prompt)} and \textit{Qwen3-0.6B (Prompt)} achieve reasonable live accuracy in the range of 58–63\%, but their results are notably weaker in certain subcategories. By contrast, the very compact \textit{TinyLlama-1.1B-32k-Instruct} and \textit{TinyAgent-1.1B} achieve only 39\% non-live execution accuracy and show near-zero success on AST subtasks such as Python or JavaScript function calls. Overall, the live evaluation highlights the gap between models above one billion parameters and ultra-lightweight variants, especially for reliable execution in diverse contexts.

\subsection{Non-Live Function Calling (AST-level Metrics)}
\begin{table*}[ht]
\centering
\caption{Non-Live Accuracy (AST-level)}
\begin{tabular}{lccccccc}
\toprule
\textbf{Metric} & \textbf{Q3-4B} & \textbf{Q3-1.7B} & \textbf{x2-3B} & \textbf{Q3-0.6B} & \textbf{x2-1B} & \textbf{TA-1.1B} & \textbf{TL-1.1B} \\
\midrule
Non\_Live Overall Acc  & 88.22\% & 82.58\% & 80.03\% & 72.42\% & 67.78\% & 20.00\% & 20.00\% \\
AST Summary            & 88.50\% & 81.88\% & 82.96\% & 70.00\% & 68.79\% & 0.00\%  & 0.00\%  \\
Simple AST             & 77.00\% & 73.00\% & 75.33\% & 65.50\% & 63.17\% & 0.00\%  & 0.00\%  \\
Python Simple AST      & 95.00\% & 89.00\% & 92.00\% & 83.50\% & 83.50\% & 0.00\%  & 0.00\%  \\
Java Simple AST        & 66.00\% & 62.00\% & 62.00\% & 49.00\% & 52.00\% & 0.00\%  & 0.00\%  \\
JavaScript Simple AST  & 70.00\% & 68.00\% & 72.00\% & 64.00\% & 54.00\% & 0.00\%  & 0.00\%  \\
Multiple AST           & 95.50\% & 89.00\% & 92.00\% & 76.00\% & 83.00\% & 0.00\%  & 0.00\%  \\
Parallel AST           & 91.00\% & 82.00\% & 86.00\% & 72.50\% & 73.00\% & 0.00\%  & 0.00\%  \\
Parallel Multiple AST  & 90.50\% & 83.50\% & 78.50\% & 66.00\% & 56.00\% & 0.00\%  & 0.00\%  \\
Irrelevance Detection  & 87.08\% & 85.42\% & 68.33\% & 82.08\% & 63.75\% & 100.00\%& 100.00\% \\
\bottomrule
\end{tabular}
\\[3pt]
\footnotesize Q3: Qwen3; x2: xLAM-2; TL: TinyLlama; TA: TinyAgent.
\label{tab:nonlive}
\end{table*}
Table~\ref{tab:nonlive} reports the non-live evaluation results, where correctness is assessed at the abstract syntax tree level rather than through execution. Here, syntactic validity is emphasized across multiple programming languages and task types. The best overall performance is obtained by \textit{xLAM-2-3b-fc-r (FC)} with 80.03\% non-live accuracy and high scores across multiple AST categories, including 92.00\% in Python and 82.96\% in AST summary. The \textit{Qwen3-4B (Prompt)} model slightly outperforms with 88.22\% overall accuracy and particularly strong performance in Python AST tasks at 95.00\%. Smaller models such as \textit{Qwen3-1.7B (Prompt)} also achieve competitive results, surpassing 80\% in many categories, while \textit{Qwen3-0.6B (Prompt)} and \textit{xLAM-2-1b-fc-r (FC)} drop into the 65–72\% range. In stark contrast, both \textit{TinyLlama-1.1B-32k-Instruct} and \textit{TinyAgent-1.1B} fail to generalize, remaining at 20\% overall with zero success in Python, Java, or JavaScript AST subcategories. Notably, these smaller models only succeed in irrelevance detection, where they abstain correctly. This reinforces that syntactic correctness benefits significantly from model scale, while very small models struggle to handle multi-language and multi-function AST tasks.

\subsection{Multi-Turn Evaluation}
\begin{table*}[ht]
\centering
\caption{Multi-Turn Accuracy}
\begin{tabular}{lccccc}
\toprule
\textbf{Model} & \textbf{Overall} & \textbf{Base} & \textbf{Miss Func} & \textbf{Miss Param} & \textbf{Long Ctx} \\
\midrule
xLAM-2-3b-fc-r (FC)    & 55.62\% & 69.50\% & 56.50\% & 55.50\% & 41.00\% \\
xLAM-2-1b-fc-r (FC)    & 35.25\% & 43.00\% & 39.00\% & 35.00\% & 24.00\% \\
Qwen3-4B (Prompt)      & 16.88\% & 21.50\% & 18.50\% & 14.00\% & 13.50\% \\
Qwen3-1.7B (Prompt)    & 8.38\%  & 10.50\% & 8.50\%  & 9.50\%  & 5.00\%  \\
Qwen3-0.6B (Prompt)    & 1.38\%  & 1.50\%  & 1.50\%  & 1.50\%  & 1.00\%  \\
TinyAgent-1.1B         & 0.00\%  & 0.00\%  & 0.00\%  & 0.00\%  & 0.00\%  \\
TinyLlama-1.1B-32k-Inst& 0.00\%  & 0.00\%  & 0.00\%  & 0.00\%  & 0.00\%  \\
\bottomrule
\end{tabular}
\label{tab:multiturn}
\end{table*}
Table~\ref{tab:multiturn} highlights multi-turn interaction performance, which is the most demanding evaluation scenario as it requires context retention across sequential exchanges. The \textit{xLAM-2-3b-fc-r (FC)} model achieves the highest score with 55.62\% overall multi-turn accuracy, supported by strong results in base cases (69.50\%) and consistent handling of missing functions and parameters. The \textit{xLAM-2-1b-fc-r (FC)} follows with 35.25\%, indicating that scaling within the same family yields significant gains for multi-turn tasks. The \textit{Qwen3-4B (Prompt)} and \textit{Qwen3-1.7B (Prompt)} models achieve 16.88\% and 8.38\% respectively, reflecting limited ability to manage long-context reasoning when constrained to prompt-based function calling. The smallest variant, \textit{Qwen3-0.6B (Prompt)}, is almost entirely unable to sustain multi-turn performance, reaching only 1.38\%. Both \textit{TinyLlama-1.1B-32k-Instruct} and \textit{TinyAgent-1.1B} record zero performance across all subcategories, failing to demonstrate any interactive reliability. These results confirm that multi-turn dialogue is particularly sensitive to model size and optimization method, with smaller SLMs incapable of retaining sufficient context for accurate function calling across turns.

\section{Discussion}
The evaluation results highlight key trade-offs between model scale, optimization, and reliability for agentic small language models (SLMs). Ultra-compact models like TinyLlama and TinyAgent underperform across all settings, with accuracies below 20\% and no multi-turn success, indicating they are unsuitable for robust agentic deployment without further improvements.

In contrast, SLMs in the 1–3B parameter range, notably xLAM and Qwen, achieve much higher accuracy—often exceeding 80\% on AST correctness and over 75\% in live execution. The xLAM-2-3B-fc-r model also excels in multi-turn tasks with over 55\% accuracy, showing that medium-scale, well-optimized SLMs can sustain context and multi-step reasoning.

Methodologically, supervised fine-tuning (SFT) offers limited gains and struggles with complex multi-function tasks. Preference-based methods like Direct Preference Optimization (DPO) improve output stability and alignment, while reinforcement learning (e.g., PPO) is effective but computationally demanding—less ideal for edge deployment. Hybrid approaches combining SFT initialization with reinforcement learning refinement balance stability and adaptability, though with higher resource costs.

Evaluation diversity is vital: while AST correctness is fundamental, execution-based and cross-language tests reveal compact models' limitations. Relevance detection and conversational robustness further emphasize the multi-dimensional nature of agent reliability. Comprehensive frameworks like BFCL integrate these facets for thorough assessment.

Overall, while ultra-compact SLMs remain limited, medium-scale models with efficient fine-tuning and preference alignment can reliably support real-world deployment on constrained devices. A layered optimization strategy—PEFT-based SFT for efficiency, preference alignment for stability, and hybrid reinforcement for adaptability—offers a practical roadmap for agentic SLMs in edge environments.

\section{Conclusion and Ongoing Work}
This study shows small language models under 3 billion parameters can effectively handle agentic tasks like tool and API calling, enabling low-latency, privacy-conscious, and cost-efficient deployment on edge devices. Using the Berkeley Function-Calling Leaderboard, we benchmarked multiple SLM families, revealing limitations of ultra-compact models and strong performance from medium-scale ones such as xLAM and Qwen across diverse task categories. We examined trade-offs among supervised fine-tuning, parameter-efficient methods, reinforcement learning, and hybrid approaches, highlighting Direct Preference Optimization with AgentBank data as a cost-effective alignment method with hybrid approaches enhancing adaptability.

We developed a DPO pipeline that converted ALFRED data into preference pairs, with ongoing work on ALFWorld to enable large-scale, preference-aligned BFCL evaluation. Our findings emphasize the importance of broad evaluation beyond syntax to include execution, relevance, and conversational integration for real-world readiness.

Future work includes scaling DPO training on varied datasets, refining hybrid fine-tuning cycles to reduce compute overhead while maintaining flexibility, expanding benchmarks to cover cross-language and multi-modal function calling, and conducting longitudinal evaluations to track progress and ensure reproducibility. Overall, this work lays a foundation and roadmap for creating efficient, reliable small language model agents suited for edge deployment.

\bibliographystyle{IEEEtran}
\bibliography{references}

\end{document}